\documentclass[11pt,a4paper]{article}
\usepackage[hyperref]{acl2021}
\usepackage{times}
\usepackage{latexsym}

\usepackage{microtype}

\usepackage{amsmath}
\usepackage{array,multirow}
\usepackage{booktabs, hhline}
\usepackage{color, colortbl}
\usepackage{enumitem}
\usepackage{graphicx}
\usepackage{siunitx}
\usepackage{subcaption}

\aclfinalcopy 

\definecolor{light-gray}{gray}{0.9}

\graphicspath{{figures/}}

\title{Attention Is Indeed All You Need:\\ Semantically Attention-Guided Decoding for Data-to-Text NLG}
  
\author{Juraj Juraska \and Marilyn Walker \\
Natural Language and Dialogue Systems Lab \\
University of California, Santa Cruz \\
\texttt{\{jjuraska,mawalker\}@ucsc.edu} \\}

\date{}

\begin{document}

\maketitle

\begin{abstract}

Ever since neural models were adopted in data-to-text language generation, they have invariably been reliant on extrinsic components to improve their semantic accuracy, because the models normally do not exhibit the ability to generate text that reliably mentions all of the information provided in the input.
In this paper, we propose a novel decoding method that extracts interpretable information from encoder-decoder models' cross-attention, and uses it to infer which attributes are mentioned in the generated text, which is subsequently used to rescore beam hypotheses. Using this decoding method with T5 and BART, we show on three datasets its ability to dramatically reduce semantic errors in the generated outputs, while maintaining their state-of-the-art quality.

\end{abstract}

\section{Introduction}


Task-oriented dialogue systems require high semantic fidelity of the generated responses in order to correctly track what information has been exchanged with the user. Therefore, their natural language generation (NLG) components are typically conditioned on structured input data, performing \emph{data-to-text} generation. To achieve high semantic accuracy, neural models for data-to-text NLG have invariably been reliant on extrinsic components or methods. While large pretrained generative language models (LMs), such as GPT-2 or T5, perform better in this respect, even they do not normally generate text that reliably mentions all the information provided in the input. 

In this work, we study the behavior of attention in large pretrained LMs fine-tuned for data-to-text NLG tasks. We show that \emph{encoder-decoder} models equipped with \emph{cross-attention} (i.e., an attention mechanism in the decoder looking back at the encoder's outputs) are, in fact, aware of the semantic constraints, yet standard decoding methods do not fully utilize the model's knowledge. The method we propose extracts interpretable information from the model's cross-attention mechanism at each decoding step, and uses it to infer which slots have been correctly realized in the output. Coupled with beam search, we use the inferred slot realizations to rescore the beam hypotheses, preferring those with the fewest missing or incorrect slot mentions.

To summarize our contributions, the proposed semantic attention-guided decoding method, or \textsc{SeA-GuiDe} for short: \textbf{(1)}~drastically reduces semantic errors in the generated text (shown on the E2E, ViGGO, and MultiWOZ datasets); \textbf{(2)}~is domain- and model-independent for encoder-decoder architectures with cross-attention, as shown on different sizes of T5 and BART; \textbf{(3)}~works out of the box, but is parameterizable, which allows for further optimization; \textbf{(4)}~adds only a small performance overhead over beam search decoding; and \textbf{(5)}~perhaps most importantly, requires no model modifications, no additional training data or data preprocessing (such as augmentation, segmentation, denoising, or alignment), and no manual annotation.\footnote{The code for \textsc{SeA-GuiDe} and heuristic semantic error evaluation can be found at \url{https://github.com/jjuraska/data2text-nlg}.}


\section{Related Work}

Several different approaches to enhancing semantic accuracy of neural end-to-end models have been proposed for data-to-text NLG over the years.
The most common approach to ensuring semantic quality relies on over-generating and then reranking candidate outputs using
criteria that the model 
was not explicitly optimized for in training. Reranking in sequence-to-sequence models is typically performed by creating an extensive set of rules, or by training a supplemental classifier, that indicates for each 
input slot whether it is present in the output utterance~\cite{wen2015stochastic,duvsek2016sequence,juraska2018deep,agarwal2018char2char,kedzie2020controllable,harkous2020have}. 

\citet{wen2015semantically} proposed an extension of the underlying LSTM cells of their sequence-to-sequence model to explicitly track, at each decoding step, the information mentioned so far. The coverage mechanism~\citep{tu2016modeling,mi2016coverage,see2017get} penalizes the model for attending to the same parts of the input based on the cumulative attention distribution in the decoder. \citet{chisholm2017learning} and \citet{shen2019pragmatically} both introduce different sequence-to-sequence model architectures that jointly learn to generate text and reconstruct the input facts. An iterative self-training process using data augmentation~\citep{nie2019simple,kedzie2019good} 
was shown to reduce semantic NLG errors on 
the E2E dataset~\citep{novikova2017e2e}. Among the most recent efforts, the jointly-learned segmentation and alignment method of~\citet{shen2020neural} improves semantic accuracy while simultaneously increasing output diversity. \citet{kedzie2020controllable} use segmentation for data augmentation and automatic utterance planning, which leads to a reduction in semantic errors on both the E2E and ViGGO~\citep{juraska2019viggo} datasets.

In contrast to the above methods, our approach does not rely on model modifications, data augmentation, or manual annotation. Our method is novel in that it utilizes information that is already present in the model itself to perform semantic reranking.


Finally, related to our work is also controllable neural language generation, in which the constrained decoding strategy is often used, rescoring tokens at each decoding step based on a set of feature discriminators~\cite{ghazvininejad2017hafez,baheti2018generating,holtzman2018learning}. Nevertheless, this method is typically used with unconditional generative LMs, and hence does not involve input-dependent constraints.

\section{Semantic Attention-Guided Decoding}
\label{sec:sea-guide}

While we will evaluate the \textsc{SeA-GuiDe} method on ViGGO, E2E, and MultiWOZ, we develop the method by careful analysis of the cross-attention behavior of different pretrained generative LMs fine-tuned on the ViGGO dataset. ViGGO is a parallel corpus of structured meaning representations~(MRs) and corresponding natural-language utterances in the video game domain. The MRs consist of a dialogue act (DA) 
and a list of slot-and-value pairs. The motivation for selecting ViGGO for developing the method was that it is the smallest dataset, but it provides a variety of DA and slot types (as shown in Table~\ref{tab:dataset_overview}). The models used for the analysis were the smallest variants of T5~\citep{raffel2020exploring} and BART~\citep{lewis2020bart}. We saved the larger variants of the models, as well as the other two datasets, for the evaluation.

\subsection{Interpreting Cross-Attention}
\label{sec:interpreting-cross-attention}

Attention~\citep{bahdanau2015neural,luong2015effective} is a mechanism that was introduced in encoder-decoder models~\citep{sutskever2014sequence,cho2014learning} to overcome the long-range dependencies problem of RNN-based models. It~allows the decoder to effectively condition its output tokens on relevant parts of the encoder's output at each decoding step. The term \emph{cross-attention} is primarily used when referring to the more recent transformer-based encoder-decoder models~\citep{vaswani2017attention}, to distinguish it from the \emph{self-attention} layers present in both the encoder and the decoder transformer blocks. The cross-attention layer ultimately provides the decoder with a weight distribution at each step, indicating the importance of each input token in the current context.

\begin{figure*}
    \centering
    \begin{subfigure}[t]{0.31\textwidth}
        \fbox{\includegraphics[width=\linewidth]{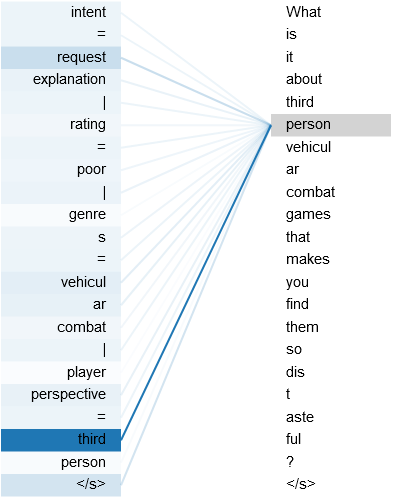}}
        \caption{Verbatim slot mention (1\textsuperscript{st} layer).}
        \label{subfig:attention-visualization-verbatim}
    \end{subfigure}
    \hfill
    \begin{subfigure}[t]{0.31\textwidth}
        \fbox{\includegraphics[width=\linewidth]{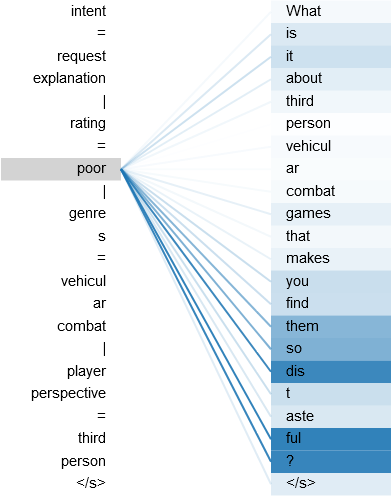}}
        \caption{Paraphrased slot mention (3\textsuperscript{rd} layer).}
        \label{subfig:attention-visualization-paraphrased}
    \end{subfigure}
    \hfill
    \begin{subfigure}[t]{0.31\textwidth}
        \fbox{\includegraphics[width=\linewidth]{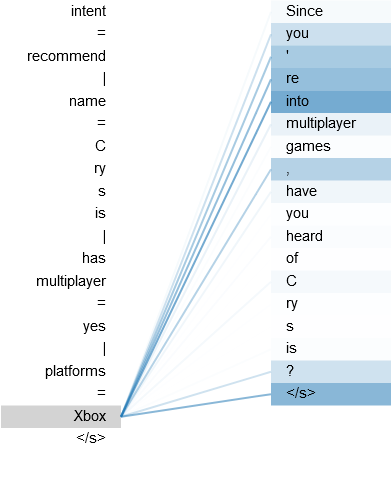}}
        \caption{Unrealized slot mention (4\textsuperscript{th} layer).}
        \label{subfig:attention-visualization-unrealized}
    \end{subfigure}
    \vspace{-0.2cm}
    \caption{Visualization of cross-attention weight distribution for the 6-layer T5-small (trained on the ViGGO dataset) in 3 different scenarios. The left column in each corresponds to the input tokens, and the right to the tokens generated by the decoder. The darker the blue background shade, the greater the attention weight. Note that the weights are aggregated across all attention heads by extracting the maximum.}
    \label{fig:attention-visualization}
    \vspace{-0.4cm}
\end{figure*}

Our results below will show that visualizing the attention weight distribution for individual cross-attention layers in the decoder -- for many different inputs -- reveals multiple universal patterns, whose combination can be exploited to track the presence, or lack thereof, of input slots in the output sequence. Despite the differences in the training objectives of T5 and BART, as well as their different sizes, we observe remarkably similar patterns in their respective cross-attention behavior. Below, we describe the three most essential patterns (illustrated in Figure~\ref{fig:attention-visualization}) that we use in \textsc{SeA-GuiDe}.

\subsubsection{Verbatim Slot Mention Pattern}
\label{sec:verbatim-slot-mention-pattern}

The first pattern  consistently occurs in the lowest attention layer, whose primary role appears to be to retrospectively keep track of a token in the input sequence that the decoder just generated in the previous step. Figure~\ref{subfig:attention-visualization-verbatim} shows an example of an extremely high attention weight on the input token ``third'' when the decoder is deciding which token to generate after ``What is it about \emph{third}'' (which ends up being the token ``person''). This pattern, which  we refer to as the \emph{verbatim} slot mention pattern, can be captured by maximizing the weight over all attention heads in the decoder's first layer.

\subsubsection{Paraphrased Slot Mention Pattern}
\label{sec:paraphrased-slot-mention-pattern}

\emph{Paraphrased} slot mentions, on the other hand, are captured by the higher layers, at the moment when a corresponding token is about to be mentioned next. Essentially, as we move further up the layers, the cross-attention weights gradually shift towards input tokens that correspond to information that is most likely to \emph{follow next} in the output, and capture increasingly more abstract concepts in general.  Figure~\ref{subfig:attention-visualization-paraphrased} shows an example of the \textsc{rating} slot's value ``poor'' paraphrased in the generated utterance as ``distasteful''; the first high attention value associated with the input token ``poor'' occurs when the decoder is about to generate the ``dis'' token.

\begin{figure}
    \centering
    \fbox{\includegraphics[width=0.59\linewidth]{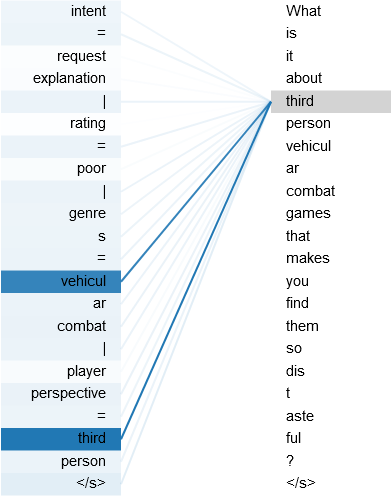}}
    \vspace{-0.1cm}
    \caption{An example of the decoder paying equal attention (in the 5\textsuperscript{th} layer of the 6-layer T5-small) to two slots in the input sequence when deciding what to generate next after ``What is it about''.}
    \label{subfig:attention-visualization-multiple-choice}
    \vspace{-0.4cm}
\end{figure}

At certain points during generation, however, the attention in the uppermost layers is distributed fairly evenly among multiple slots, because any of them could lead to a coherent continuation of the sentence.
For example, the generated utterance in Figure~\ref{subfig:attention-visualization-multiple-choice} could have  started with ``What is it about vehicular combat games played from a third-person perspective that\ldots'', where the \textsc{genres} slot is output before the \textsc{player perspective} slot.

In order to recognize a paraphrased mention, without incorrectly capturing other slots considered, we propose averaging the cross-attention weights, using only the bottom half of the layers (e.g., layers 1 to 3 in the T5-small model).

\subsubsection{Unrealized Slot Mention Pattern}

The third pattern alleviates any undesired side effects of identifying paraphrased mentions using the second pattern, i.e., slots incorrectly assumed to be mentioned. Figure~\ref{subfig:attention-visualization-unrealized} illustrates  an unrealized slot (\textsc{platforms}) being paid attention to in several decoding steps.  The cross-attention weight distribution for the ``Xbox'' token in the 4th layer, shows that the decoder considered mentioning the slot at step~5 (e.g., ``Since you're \emph{an Xbox fan} and like multiplayer games,\ldots''), as well as step~8 (e.g., ``\ldots into multiplayer games \emph{on Xbox,}\ldots''). The second pattern, depending on the sensitivity setting (see Section~\ref{sec:slot-mention-tracking}), might infer the \textsc{platforms} slot as  a paraphrased mention at step 5 and/or 8.

However, the \textsc{platforms} slot's value is also paid attention to when the decoder is about to generate the EOS token and, importantly, without any high attention weights associated with other slots at this step. This suggests that the model \emph{is aware} that it omitted that slot. However, at that point, the decoder is more confident ending the sentence than realizing the missed slot after generating a question mark. This \emph{unrealized} slot mention pattern is most likely to occur in the higher cross-attention layers, but not necessarily, so it is more effective to capture it by averaging the attention weights over all layers (at the last decoding step).

\paragraph{Note on Boolean Slots.}
With any of the three patterns described above, Boolean slots, such as \textsc{has multiplayer} in Figure~\ref{subfig:attention-visualization-unrealized}, typically have a high attention weight associated with their name rather than the value. This observation leads to a different treatment of Boolean slots, as described in Appendix~\ref{sec:appendix-slot-mention-tracking}.

\subsection{Slot Mention Tracking}
\label{sec:slot-mention-tracking}

We use the findings of the cross-attention analysis for automatic slot mention tracking in the decoder. During decoding, for each sequence, the attention weights associated with the next token to be generated are aggregated as per Section~\ref{sec:interpreting-cross-attention}. Using configurable \emph{thresholds}, the aggregated weights are then binarized, i.e., set to~$1$ if above the threshold, and $0$ otherwise. This determines the sensitivity of the pattern recognition. Optionally, all but the maximum weight can be set to~$0$, in which case only a single input token will by implied even if the attention mass is spread evenly across multiple tokens. Finally, the indices of binarized weights of value $1$, if any, are matched with their corresponding slots depending on which slot-span in the input sequence they fall into. For details on automatically extracting slot spans, see  Appendix~\ref{sec:appendix-slot-mention-tracking}.

\subsubsection{Mention-Tracking Components}
\label{sec:mention-tracking-components}

The three mention-tracking components, each of which operates on different attention layers and uses a different weight aggregation and binarization strategy, are summarized in Table~\ref{tab:mention-tracking-components}. These components are executed in sequence and update one common slot-tracking object.

The first component, which tracks verbatim mentions, operates on the first attention layer only, with a high binarization threshold. Slot mentions identified by this component are regarded as high-confidence. The second component tracks paraphrased mentions, which are identified as slot mentions with low confidence, due to the partial ambiguity in mention detection using the second pattern (see Section~\ref{sec:paraphrased-slot-mention-pattern}). The third component only kicks in when the EOS token is the most probable next token. At that point, it identifies -- with high sensitivity -- slots that were not realized in the sequence (e.g., the \textsc{platforms} slot in Figure~\ref{subfig:attention-visualization-unrealized}), and removes the corresponding mention record(s). Only low-confidence mentions can be erased, while high-confidence ones are final once they are detected.

\subsection{Semantic Reranking}

Combining the slot mention tracking with beam search, for each input MR we obtain a pool of candidate utterances along with the semantic errors inferred at decoding time. We then rerank the candidates and pick the one with the fewest errors, resolving ties using the length-weighted log-probability scores determined during beam search.


\section{Evaluation}

In order to measure the proposed decoding method's performance in semantic error reduction, we first develop an \emph{automatic} way of identifying erroneous slot mentions in generated utterances. In a human evaluation we establish that its performance is nearly perfect for all three datasets used for testing our models (see Section~\ref{sec:automatic-slot-error-evaluation}). We then use it to calculate the slot error rate (SER) automatically for all our model outputs across all datasets and configurations tested, which would be infeasible to have human annotators do.

\paragraph{Datasets.}

Besides ViGGO, which we use for fine-tuning the decoding (slot-tracking) parameters of the proposed \textsc{SeA-GuiDe} method, we evaluate its effectiveness for semantic error reduction on two \emph{unseen} and \emph{out-of-domain} datasets. While E2E~\citep{novikova2017e2e} is also a simple MR-to-text generation dataset (in the restaurant domain), MultiWOZ~2.1~\citep{eric2020multiwoz} is a dialogic corpus covering several domains from which we extract system turns only, along with their MR annotations, along the lines of~\citet{peng2020few} and \citet{kale2020text}. Table~\ref{tab:dataset_overview} gives an overview of the datasets' properties.

\begin{table}
    \small
   	\centering
    \begin{tabular}{>{\centering\arraybackslash} m{0.2\linewidth} >{\centering\arraybackslash} m{0.15\linewidth} >{\centering\arraybackslash} m{0.15\linewidth} >{\centering\arraybackslash} m{0.11\linewidth} >{\centering\arraybackslash} m{0.11\linewidth}}
    	\toprule
    	& \textbf{Size}    & \textbf{Domains}   & \textbf{DAs}  & \textbf{Slots} \\
    	\midrule
    	\textbf{ViGGO}    & 6,900    & 1 & 9   & 14 \\
    	\textbf{E2E}    & 51,426    & 1 & 1  & 8 \\
    	\textbf{MultiWOZ}    & 70,530    & 7 & 13 & 27 \\
        \bottomrule
    \end{tabular}
    \vspace{-0.2cm}
	\caption{Dataset statistics, including the total number of dialogue act (DA) and slot types. For MultiWOZ, the numbers are calculated across system turns only.}
    \label{tab:dataset_overview}
    \vspace{-0.2cm}
\end{table}

\paragraph{Setup.}

In our experiments, we fine-tune T5 and BART models of varying sizes on the above datasets' training partitions, select the best model checkpoints based on the BLEU score they achieve on the respective validation set, and evaluate them on the test sets while using different decoding methods for inference. For beam search decoding, including when used as part of \textsc{SeA-GuiDe}, we use beam size 10 and early stopping, unless stated otherwise. All of our results are averaged over three runs with random initialization. For further details on training and inference parameters, we refer the reader to Appendix~\ref{sec:appendix-model-parameters}.


\subsection{Automatic Slot Error Evaluation}
\label{sec:automatic-slot-error-evaluation}

We evaluate our trained models performance with the standard NLG metrics BLEU~\cite{papineni2002bleu}, METEOR~\cite{lavie2007meteor}, ROUGE-L~\cite{lin2004rouge}, and CIDEr~\cite{vedantam2015cider}, whose calculation is detailed in Appendix~\ref{sec:appendix-evaluation-metrics}.
However, we also put substantial effort into developing a highly accurate heuristic \emph{slot aligner} to calculate the semantic accuracy of generated utterances.
The slot aligner is rule-based and took dozens of man-hours to develop, but it is robust and extensible to new domains, so it works on all three test datasets. Using the slot aligner, we count missed, incorrect, and repeated slot mentions, and determine the slot error rate (SER) as the percentage of these errors out of all slots.

\begin{table}
    \small
   	\centering
    \begin{tabular}{>{\centering\arraybackslash} m{0.16\linewidth} >{\centering\arraybackslash} m{0.09\linewidth} >{\centering\arraybackslash} m{0.26\linewidth} >{\centering\arraybackslash} m{0.14\linewidth} >{\centering\arraybackslash} m{0.07\linewidth}}
    	\toprule
    	& \textbf{SER\textsubscript{SA}}  & \textbf{SER CI (95\%)}    & \textbf{Precision}  & \textbf{IAA} \\
    	\midrule
    	\textbf{ViGGO}  & $2.77\%$  & $2.19 \pm 1.55\%$ & $97.37\%$    & $1.00$ \\
    	\textbf{E2E}    & $3.98\%$  & $3.91 \pm 1.73\%$ & $100\%$    & $1.00$ \\
    	\textbf{MultiWOZ}    & $1.19\%$    & $1.35 \pm 0.91\%$ & $94.89\%$  & $0.90$ \\
        \bottomrule
    \end{tabular}
    \vspace{-0.2cm}
	\caption{Human evaluation of the slot aligner's performance on each dataset. The IAA column indicates the Krippendorff's alpha reliability coefficient.}
    \label{tab:slot_aligner_performance}
    \vspace{-0.4cm}
\end{table}

To verify our slot aligner's performance, we take the generated utterances of one model per dataset for which it calculated a relatively high SER (indicated in the SER\textsubscript{SA} column in Table~\ref{tab:slot_aligner_performance}). We then have one of the authors and an additional expert annotator manually label all of the errors as true or false positives. This corresponds to 38, 173 and 176 errors for ViGGO, E2E and MultiWOZ, respectively. From that we calculate the precision for each dataset, which turns out to be above 94\% for each of the datasets. The almost perfect inter-annotator agreement (IAA), besides validating the precision, also suggests that the SER is an objective metric, and therefore well-suited for automation.

Furthermore, we take samples of $72$ ($\approx 20\%$), $63$ ($\approx 10\%$) and $290$ ($\approx 4\%$) of the generated utterances on ViGGO, E2E and MultiWOZ, respectively, annotate them for all types of errors, and calculate the \emph{actual} SER confidence intervals (middle column). Their good alignment with the slot aligner SER scores, together with the high error classification precision, leads us to the conclusion that the slot aligner performs similarly to humans in identifying semantic errors on the above datasets.

Besides SER evaluation, the slot aligner can also be used for beam reranking.
Due to the handcrafted and domain-specific nature of the slot aligner, beam search with this reranking has a distinct advantage over \textsc{SeA-GuiDe}, which can be used for any domain out of the box.
We therefore consider the results when using the slot-aligner reranking to be an upper bound for \textsc{SeA-GuiDe} in terms of SER.

\subsection{\textsc{SeA-GuiDe} Parameter Tuning}

Each of the three mention-tracking components described in Section~\ref{sec:mention-tracking-components} has four configurable parameters, which we tuned by testing T5-small and BART-base, fine-tuned on the ViGGO dataset and equipped with \textsc{SeA-GuiDe} for inference. The parameter optimization was based on the insights obtained in Section~\ref{sec:interpreting-cross-attention} and a subsequent grid search, with results in Table~\ref{tab:mention-tracking-components}.

\begin{table}
    \small
   	\centering
    \begin{tabular}{>{\centering\arraybackslash} m{0.19\linewidth} >{\centering\arraybackslash} m{0.15\linewidth} >{\centering\arraybackslash} m{0.28\linewidth} >{\centering\arraybackslash} m{0.16\linewidth}}
    	\toprule
    	& \textbf{Verbatim}   & \textbf{Paraphrased}    & \textbf{Unrealized} \\
    	\midrule
    	\textbf{Layer agg.}    & 1\textsuperscript{st} layer only & avg. over bottom half of layers    & avg. \\
    	\midrule
    	\textbf{Head agg.}    & max. & max.    & max. \\
    	\midrule
    	\textbf{Bin. threshold}    & 0.9 & 0.4 (T5-small) 0.3 (BART-base)    & 0.1 \\
    	\midrule
    	\textbf{Bin. max.}    & yes    & no & no \\
        \bottomrule
    \end{tabular}
    \vspace{-0.2cm}
	\caption{The final configuration of parameters used in each of the 3 mention-tracking components. The ``Bin. max.'' row indicates whether only the maximum weight is kept during binarization, or all above the threshold.}
    \label{tab:mention-tracking-components}
    \vspace{-0.4cm}
\end{table}

\begin{figure}
    \centering
    \begin{subfigure}[t]{\linewidth}
        \includegraphics[width=\linewidth]{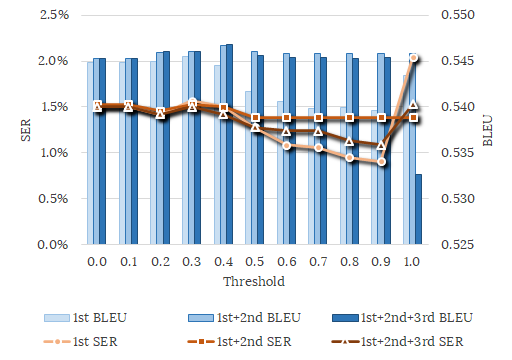}
        \caption{Threshold optimization for the 1\textsuperscript{st}~component (verbatim mentions), with the other components enabled or disabled. When enabled, the 2\textsuperscript{nd}~component's threshold was fixed at $0.3$, and that of the 3\textsuperscript{rd} at $0.1$. Note that the threshold of $1.0$ is equivalent to the 1\textsuperscript{st}~component being disabled, as attention weights are in the $[0.0, 1.0]$ range.}
        \label{subfig:thresholds-1st-component}
        \vspace{0.1in}
    \end{subfigure}
    \begin{subfigure}[t]{\linewidth}
        \includegraphics[width=\linewidth]{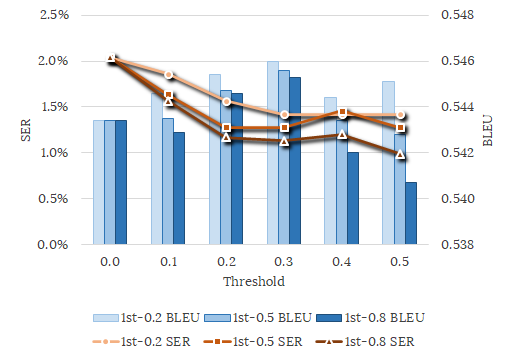}
        \caption{Threshold optimization for the 2\textsuperscript{nd}~component (paraphrased mentions), with the 1\textsuperscript{st}~component's threshold of $0.2$, $0.5$ and $0.8$, and that of the 3\textsuperscript{rd}~component fixed at $0.1$.}
        \label{subfig:thresholds-2nd-component}
        \vspace{0.1in}
    \end{subfigure}
    \begin{subfigure}[t]{\linewidth}
        \includegraphics[width=\linewidth]{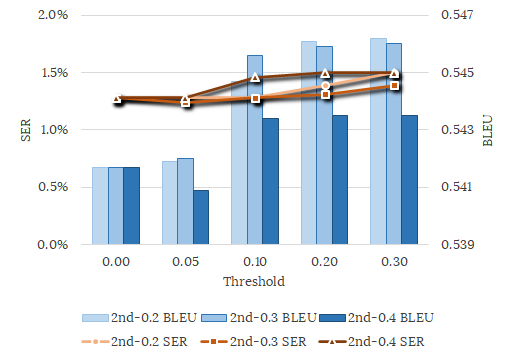}
        \caption{Threshold optimization for the 3\textsuperscript{rd}~component (unrealized mentions), with the 2\textsuperscript{nd}~component's threshold of $0.2$, $0.3$ and $0.4$, and that of the 1\textsuperscript{st}~component fixed at $0.5$.}
        \label{subfig:thresholds-3rd-component}
    \end{subfigure}
    \caption{Effects of different parameter configurations of the 3 mention-tracking components on SER and BLEU of utterances generated by BART-base fine-tuned on ViGGO.}
    \label{fig:binarization-threshold-optimization}
\end{figure}

For attention weight aggregation, we experimented with summing, averaging, maximizing, and normalizing. We determined \emph{averaging} over layers and \emph{maximizing} over heads to be the best combination for all three components. As for the binarization thresholds, Figure~\ref{fig:binarization-threshold-optimization} shows the most relevant slice of the grid search space for each component, leading to the final threshold values.

To show the effect of each slot-tracking component, we perform an ablation study with individual components disabled.\footnote{The 3\textsuperscript{rd} component has no effect without the 2\textsuperscript{nd}, so we do not consider the combination where only the 2\textsuperscript{nd} is disabled.} As the plot in Figure~\ref{subfig:thresholds-1st-component} demonstrates, the 1\textsuperscript{st} component by itself reduces the SER the most, but at the expense of the BLEU score, which decreases as the SER does -- to the point where BLEU drops below $0.54$ when the SER is at its lowest ($0.91\%$), that is with a threshold of $0.9$. For reference, the SER and the BLEU score achieved with beam search only are $2.04\%$ and $0.543$, respectively. Adding the 2\textsuperscript{nd} component brings the BLEU score up to above $0.545$, nevertheless the SER jumps to $1.39\%$. Finally, enabling the 3\textsuperscript{rd} component too has a negligible negative effect on BLEU, but reduces the SER to $1.09\%$.

Figure~\ref{subfig:thresholds-2nd-component} shows that the 2\textsuperscript{nd} component gives optimal performance when its threshold is set to around $0.3$. This setting maximizes BLEU, while keeping SER low. Beyond $0.3$ the BLEU score starts dropping fast, and with a threshold of greater than $0.5$, the 2\textsuperscript{nd} component has barely any effect anymore. Similarly, Figure~\ref{subfig:thresholds-3rd-component} shows the threshold value of $0.1$ to be optimal in the 3\textsuperscript{rd} component, when optimizing for both metrics. Thresholds higher than $0.3$ cut off almost all aggregated weights in this component, virtually disabling it.

\subsection{Effects of Beam Size on \textsc{SeA-GuiDe}}

\begin{figure}
    \centering
    \includegraphics[width=0.9\linewidth]{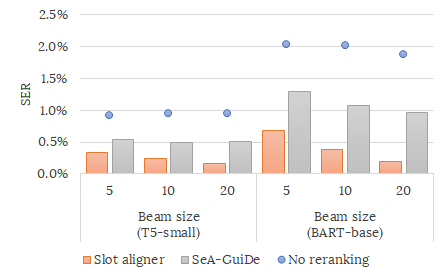}
    \vspace{-0.1in}
    \caption{The effect of different beam size on the SER using different reranking methods on the ViGGO dataset. With greedy search decoding, the SER is $1.65\%$ and $2.70\%$ for T5 and BART, respectively.}
    \label{fig:beam_search_viggo}
    \vspace{-0.4cm}
\end{figure}

Since \textsc{SeA-GuiDe} uses beam search to generate the pool of candidates that it later reranks, we analyzed the effect of increasing the beam size on the SER of the final utterances. As Figure~\ref{fig:beam_search_viggo} shows for the ViGGO dataset, \textsc{SeA-GuiDe} certainly benefits from increasing the beam size from 5 to 10, but the benefit shrinks substantially (or disappears entirely, in case of T5-small) when further increased to 20. An analysis for the E2E dataset, with similar results, is presented in Appendix~\ref{sec:appendix-beam-size-e2e}.

\section{Results}

To maximize the performance of the models using \textsc{SeA-GuiDe}, the binarization thresholds (and possibly other parameters of the mention-tracking components) can be optimized for each model and dataset on the validation set. In our evaluation, however, we focused on demonstrating the effectiveness of this decoding method out of the box. That being said, even common decoding methods, such as simple beam search or nucleus sampling~\cite{holtzman2019curious}, usually benefit from parameter optimization (e.g., beam size, or the $p$-value) whenever used with a different model or dataset.

\subsection{\textsc{SeA-GuiDe} Performance}

While developing the \textsc{SeA-GuiDe} method we analyzed the behavior of cross-attention on both the T5-small and the BART-base model; interestingly, the decoding performs best for both with nearly the same configuration. The only difference is the 2\textsuperscript{nd} component's binarization threshold (see Table~\ref{tab:mention-tracking-components}), accounting for the fact that BART-base has 50\% more attention heads than T5-small, which causes the attention weights to be more spread out.

The upper half of Table~\ref{tab:results_viggo} compares the two models' performance with \textsc{SeA-GuiDe} vs. other decoding methods, as well as against three state-of-the-art baselines. As the results show, both models, when using \textsc{SeA-GuiDe}, significantly reduce the number of semantic errors in the generated outputs compared to using greedy search ($\approx$ 3.4 and 2.5 times in case of T5 and BART, respectively) or simple beam search ($\approx$ 1.9 times both). As expected, the slot-aligner (SA) reranking achieves even better results thanks to the handcrafted rules it relies on. In addition, the overall high automatic metric scores suggest that the fluency of utterances generated using \textsc{SeA-GuiDe} does not suffer.

\begin{table}[h]
    \small
    \centering
        \begin{tabular} { >{\centering\arraybackslash}m{0.1cm} >{\centering\arraybackslash}m{0.3cm} >{\centering\arraybackslash}m{0.9cm} >{\centering\arraybackslash}m{0.8cm} >{\centering\arraybackslash}m{0.9cm} >{\centering\arraybackslash}m{0.8cm} >{\centering\arraybackslash}m{0.9cm}
        }
        \toprule
        \multicolumn{2}{c}{\textbf{Model}} & \textbf{BLEU}
        & \textbf{MET.}
        & \textbf{ROUGE}
        & \textbf{CIDEr}
        & \textbf{SER} $\downarrow$ \\
        \midrule\midrule
        \multicolumn{2}{c}{S2S}	& 0.519   & 0.388  & 0.631  & 2.531  & 2.55\% \\
        \multicolumn{2}{c}{DT}	& \textbf{0.536}   & \textbf{0.394}  & \textbf{0.640}  & \textbf{2.700}  & 1.68\% \\
        \multicolumn{2}{c}{K\&M}	& 0.485   & 0.380  & 0.592  & 2.454  & \emph{\textbf{0.46\%}} \\
        \midrule\midrule
        \multirow{4}{*}{\rotatebox[origin=c]{90}{\scriptsize T5-small}} & GS	& 0.519   & 0.387  & 0.631  & 2.647  & 1.65\% \\
        & BS	& 0.540   & 0.392  & 0.636  & 2.685  & 0.95\% \\
        & SA	& \textbf{0.541}   & \textbf{0.393}  & \textbf{0.637}  & \textbf{2.695}  & \textbf{0.24\%} \\
        \cmidrule{2-7}
        & SG	& \textbf{0.541}   & \textbf{0.393}  & \textbf{0.637}  & \textbf{2.695}  & 0.49\% \\
        \midrule
        \multirow{4}{*}{\rotatebox[origin=c]{90}{\scriptsize BART-base}} & GS	& 0.524   & 0.386  & 0.635  & 2.629  & 2.70\% \\
        & BS	& 0.544   & 0.393  & \textbf{0.639}  & 2.679  & 2.02\% \\
        & SA	& \textbf{0.547}   & \textbf{0.394}  & \textbf{0.639}  & \textbf{2.704}  & \textbf{0.39\%} \\
        \cmidrule{2-7}
        & SG	& 0.545   & 0.393  & \textbf{0.639}  & 2.698  & 1.07\% \\
        \midrule
        \multirow{4}{*}{\rotatebox[origin=c]{90}{\scriptsize T5-base}} & GS	& 0.527   & \textbf{0.394}  & \textbf{0.639}  & \textbf{2.682}  & 0.61\% \\
        & BS	& 0.534   & \textbf{0.394}  & 0.636  & 2.664  & 0.66\% \\
        & SA	& \textbf{0.536}   & \textbf{0.394}  & 0.637  & 2.672  & \textbf{0.19\%} \\
        \cmidrule{2-7}
        & SG	& \textbf{0.536}   & \textbf{0.394}  & 0.637  & 2.670  & 0.46\% \\
        \midrule
        \multirow{4}{*}{\rotatebox[origin=c]{90}{\scriptsize BART-large}} & GS	& 0.508   & 0.378  & 0.616  & 2.452  & 5.50\% \\
        & BS	& 0.535   & 0.391  & 0.628  & 2.612  & 1.78\% \\
        & SA	& \textbf{0.538}   & \textbf{0.394}  & \textbf{0.631}  & \textbf{2.659}  & \textbf{0.27\%} \\
        \cmidrule{2-7}
        & SG	& 0.533   & 0.391  & 0.627  & 2.613  & 1.41\% \\
        \bottomrule
    \end{tabular}
    \vspace{-0.2cm}
    \caption{Models tested on the ViGGO dataset using different decoding methods: greedy search (GS), beam search with no reranking (BS), beam search with slot-aligner reranking (SA), and \textsc{SeA-GuiDe} (SG). Baselines compared against are Slug2Slug~\cite{juraska2019viggo} (S2S), DataTuner~\cite{harkous2020have} (DT), and \citet{kedzie2020controllable} (K\&M). The best results are highlighted in bold for each model. SER scores of baselines reported by the authors themselves, rather than calculated using our slot aligner, are highlighted in italics, and they do not correspond exactly to our SER results.}
    \label{tab:results_viggo}
    \vspace{-0.4cm}
\end{table}

Finally, compared to the baseline models, T5-small performs on par with the state-of-the-art DataTuner in terms of automatic metrics, yet maintains a 3.4-times lower SER. This corresponds approximately to K\&M baseline's SER, whose automatic metrics, however, are significantly worse. BART-base outperforms T5-small according to most metrics, but its SER is more than double.

\subsection{Cross-Model Robustness}

In addition to T5-small and BART-base, we fine-tune a larger variant of each of the models, namely, T5-base and BART-large (see Appendix~\ref{sec:appendix-model-parameters} for model specifications), on the ViGGO dataset, and evaluate their inference performance when equipped with \textsc{SeA-GuiDe}. We do not perform any further tuning of the decoding parameters for these two models, only slightly lower the binarization thresholds (as we did for BART-base) to account for the models having more attention heads and layers. The thresholds we use for the 2\textsuperscript{nd} and 3\textsuperscript{rd} components are $\langle 0.3, 0.1 \rangle$ and $\langle 0.2, 0.05 \rangle$ for T5-base and BART-large, respectively.

The results in the lower half of Table~\ref{tab:results_viggo} show that these two larger models, fine-tuned on ViGGO, benefit from \textsc{SeA-GuiDe} beyond just the effect of beam search. T5-base performs significantly better across the board than its smaller T5 variant, so there is less room for improvement to begin with. In fact, the SER using greedy search is so low ($0.61\%$, in contrast to T5-small's $1.65\%$) that beam search causes it to increase. Nevertheless, \textsc{SeA-GuiDe} improves on both, while slightly boosting the other automatic metrics as well.

The almost twice-as-large BART-large model performs rather poorly in our experiments, in fact, significantly underperforming its smaller variant.\footnote{We observed that it frequently misrepresents names, such as ``Transportal Tycoon'' instead of ``Transport Tycoon'', which we think may be the consequence of the extremely small size of the ViGGO training set relative to the model's size.} We therefore refrain from drawing any conclusions for this model, although \textsc{SeA-GuiDe} offers a definite improvement in SER over simple beam search.

\subsection{Domain Transferability}

\begin{table}
    \small
    \centering
        \begin{tabular} { >{\centering\arraybackslash}m{0.1cm} >{\centering\arraybackslash}m{0.2cm} >{\centering\arraybackslash}m{0.9cm} >{\centering\arraybackslash}m{0.8cm} >{\centering\arraybackslash}m{0.9cm} >{\centering\arraybackslash}m{0.8cm} >{\centering\arraybackslash}m{0.9cm}
        }
        \toprule
        \multicolumn{2}{c}{\textbf{Model}} & \textbf{BLEU}
        & \textbf{MET.}
        & \textbf{ROUGE}
        & \textbf{CIDEr}
        & \textbf{SER} $\downarrow$ \\
        \midrule\midrule
        \multicolumn{2}{c}{S2S}	& 0.662   & 0.445  & 0.677  & 2.262  & 0.91\% \\
        \multicolumn{2}{c}{S$_\text{1}^\text{R}$}	& \textbf{0.686}   & \textbf{0.453}  & \textbf{0.708}  & \textbf{2.370}  & N/A \\
        \multicolumn{2}{c}{K\&M}	& 0.663   & \textbf{0.453}  & 0.693  & 2.308  & \emph{\textbf{0.00\%}} \\
        \midrule\midrule
        \multirow{4}{*}{\rotatebox[origin=c]{90}{\scriptsize T5-small}} & GS	& 0.670   & \textbf{0.454}  & 0.692  & 2.244  & 1.60\% \\
        & BS	& 0.667   & 0.453  & \textbf{0.694}  & \textbf{2.361}  & 2.85\% \\
        & SA	& \textbf{0.675}   & 0.453  & 0.690  & 2.341  & \textbf{0.02\%} \\
        \cmidrule{2-7}
        & SG	& 0.675   & 0.453  & 0.690  & 2.340  & 0.04\% \\
        \midrule
        \multirow{4}{*}{\rotatebox[origin=c]{90}{\scriptsize BART-base}}    & GS	& 0.667   & \textbf{0.454}  & 0.694  & 2.276  & 1.97\% \\
        & BS	& 0.670   & \textbf{0.454}  & \textbf{0.701}  & \textbf{2.372}  & 3.39\% \\
        & SA	& \textbf{0.680}   & 0.453  & 0.695  & 2.350  & \textbf{0.02\%} \\
        \cmidrule{2-7}
        & SG	& \textbf{0.680}   & 0.453  & 0.695  & 2.347  & 0.08\% \\
        \midrule
        \multirow{4}{*}{\rotatebox[origin=c]{90}{\scriptsize T5-base}} & GS	& 0.668   & \textbf{0.459}  & 0.692  & 2.282  & 1.85\% \\
        & BS	& 0.667   & 0.453  & \textbf{0.697}  & \textbf{2.387}  & 3.94\% \\
        & SA	& \textbf{0.682}   & 0.454  & 0.691  & 2.375  & \textbf{0.03\%} \\
        \cmidrule{2-7}
        & SG	& \textbf{0.682}   & 0.454  & 0.691  & 2.374  & 0.05\% \\
        \bottomrule
    \end{tabular}
    \vspace{-0.2cm}
    \caption{Models tested on the E2E dataset, compared against the following baselines: Slug2Slug~\cite{juraska2018deep}, (S2S) S$_\text{1}^\text{R}$~\cite{shen2019pragmatically}, and \citet{kedzie2020controllable} (K\&M).}
    \label{tab:results_e2e}
    \vspace{-0.1cm}
\end{table}

\begin{table}
    \small
    \centering
        \begin{tabular} { >{\centering\arraybackslash}m{0.1cm} >{\centering\arraybackslash}m{0.2cm} >{\centering\arraybackslash}m{0.8cm} >{\centering\arraybackslash}m{0.8cm} >{\centering\arraybackslash}m{0.7cm} >{\centering\arraybackslash}m{0.9cm} >{\centering\arraybackslash}m{1.1cm}
        }
        \toprule
        \multicolumn{2}{c}{\textbf{Model}} & \textbf{BLEU}
        & \textbf{BLEU\textsubscript{R}}
        & \textbf{MET.}
        & \textbf{SER} $\downarrow$
        & \textbf{SER\textsubscript{E}} $\downarrow$ \\
        \midrule\midrule
        \multicolumn{2}{c}{SCG}	& N/A   & 0.308  & N/A  & \emph{\textbf{0.53\%}}  & N/A \\
        \multicolumn{2}{c}{K\&R}	& N/A   & \textbf{0.351}    & N/A  & N/A  & 1.27\% \\
        \midrule\midrule
        \multirow{4}{*}{\rotatebox[origin=c]{90}{\scriptsize T5-small}} & GS	& \textbf{0.367}   & \textbf{0.351}  & \textbf{0.325}  & 1.15\%  & 1.36\% \\
        & BS	& 0.359   & 0.344  & 0.323  & 1.06\%  & 1.19\% \\
        & SA	& 0.360   & 0.344  & 0.323  & \textbf{0.41\%}  & \textbf{0.63\%} \\
        \cmidrule{2-7}
        & SG	& 0.360   & 0.344  & 0.323  & 0.60\%  & 0.85\% \\
        \midrule
        \multirow{4}{*}{\rotatebox[origin=c]{90}{\scriptsize BART-base}} & GS	& \textbf{0.372}   & \textbf{0.356}  & \textbf{0.326}  & 1.18\%  & 1.17\% \\
        & BS	& 0.363   & 0.346  & 0.323  & 1.12\%  & 1.02\% \\
        & SA	& 0.364   & 0.347  & 0.324  & \textbf{0.40\%}  & \textbf{0.60\%} \\
        \cmidrule{2-7}
        & SG	& 0.363   & 0.347  & 0.323  & 0.63\%  & 0.72\% \\
        \bottomrule
    \end{tabular}
    \vspace{-0.2cm}
    \caption{Models tested on MultiWOZ, compared against the following baselines: SC-GPT~\cite{peng2020few} (SCG) and \citet{kale2020text} (K\&R).}
    \label{tab:results_multiwoz}
    \vspace{-0.3cm}
\end{table}

We achieve similar results when evaluating across domains. Table~\ref{tab:results_e2e} shows that using \textsc{SeA-GuiDe} with all three models fine-tuned on E2E reduces the SER down to almost zero, with performance for the other metrics  comparable to the state-of-the-art baseline.\footnote{We were unable to successfully train BART-large on E2E due to the memory limitations of our computational resources.} In fact, \textsc{SeA-GuiDe} is nearly as effective at reducing errors in this dataset as the heuristic slot aligner (SA). Table~\ref{tab:results_multiwoz} compares our models against two recent baselines on the MultiWOZ dataset, where the effectiveness of \textsc{SeA-GuiDe} on SER reduction is comparable to that on the ViGGO dataset. All in all, on both the E2E and the MultiWOZ dataset, our models equipped with \textsc{SeA-GuiDe} for inference perform similarly to the best baselines for both SER and the other metrics \emph{at the same time}, whereas the baselines individually perform well according to one at the expense of the other.

\subsection{Slot Error Detection Examples}

Table~\ref{tab:error-detection-examples} shows several utterances generated for corresponding input MRs in the video game domain, along with the errors \textsc{SeA-GuiDe} detected, if any. In the first example, all slots are correctly mentioned, and \textsc{SeA-GuiDe} agrees. This utterance was ultimately selected during reranking over the beam search's choice, ``The Room is an excellent first person point-and-click puzzle game.'', which has one of the genres omitted.

The second example again showcases a successful identification of all slot mentions by \textsc{SeA-GuiDe}, this time in an utterance where our heuristic slot aligner incorrectly found an error in the \textsc{has\_lin\-ux\_re\-lease} slot mention (presumably confused by the negation shortly before the keyword ``Linux'').

The third example shows that \textsc{SeA-GuiDe}  is not confused by the difference between ``excellent'' (input slot value) and ``amazing'' (the corresponding slot realization), and it correctly identifies that the \textsc{player\_perspective} slot is not mentioned.

In the fourth example, \textsc{SeA-GuiDe} fails to recognize the incorrect mention of the \textsc{has\_lin\-ux\_re\-lease} slot. This is the most common mistake our decoding method makes.

The final example does not contain any errors, yet the heuristic slot aligner was confused by the negation formulation (as ``I wish it was'') and identified the \textsc{available\_on\_steam} slot as incorrect, while \textsc{SeA-GuiDe} did not. This demonstrates the robustness of \textsc{SeA-GuiDe} beyond exact matching. Yet despite the individual slots being mentioned semantically correctly, the generated utterance as a whole does not sound quite right. A better utterance would be something along the lines of ``Mirror's Edge Catalyst is a disappointment, I really wish they released it on Steam as well.''.

\begin{table}
    \small
   	\centering
   	\def\arraystretch{1.1}
    \begin{tabular}{p{0.94\linewidth}}
        \hline
    	\rowcolor{light-gray}
    	give\_opinion(\textsc{name}[The Room], \textsc{rating}[excellent], \textsc{genres}[indie, point-and-click, puzzle], \textsc{play\-er\_per\-spec\-tive}[first person]) \\
        \hline
    	\emph{The Room is an excellent indie point-and-click puzzle game. The first person perspective makes it easy to play.} \\ \textbf{Errors:} N/A \\
    	\hline
    	\rowcolor{light-gray}
    	verify\_attribute(name[Small World 2], \textsc{rat\-ing}[poor], \textsc{has\_lin\-ux\_release}[yes]) \\
        \hline
    	\emph{You said earlier that you disliked Small World 2. Are you generally not fond of Linux games?} \\ \textbf{Errors:} N/A \\
    	\hline
    	\rowcolor{light-gray}
    	request\_explanation(\textsc{rating} [excellent], \textsc{genres} [adventure, hack-and-slash, indie], \textsc{play\-er\_per\-spec\-tive} [third person]) \\
        \hline
    	\emph{What is it about indie hack-and-slash adventure games that you find so amazing?} \\ \textbf{Errors:} \textsc{player\_perspective} \\
        \hline
    	\rowcolor{light-gray}
    	inform(\textsc{name}[RollerCoaster Tycoon], \textsc{rating}[good], \textsc{gen\-res}[simulation, strategy], \textsc{platforms}[Xbox, PC], \textsc{available\_on\_steam}[no], \textsc{has\_linux\_release}[yes], \textsc{has\_mac\_release}[no]) \\
        \hline
    	\emph{RollerCoaster Tycoon is a good simulation strategy game for Xbox and PC. It is not available on Steam, Linux, or Mac.} \\ \textbf{Errors:} N/A (incorrect) \\
        \hline
    	\rowcolor{light-gray}
    	give\_opinion(\textsc{name} [Mirror's Edge Catalyst], \textsc{rating} [poor], \textsc{available\_on\_steam} [no]) \\
        \hline
    	\emph{Mirror's Edge Catalyst is a pretty bad game, I really wish it was available on Steam.} \\ \textbf{Errors:} N/A \\
        \hline
    \end{tabular}
 	\vspace{-0.2cm}
	\caption{Examples of slot-mention tracking results using \textsc{SeA-GuiDe}. Each gray row is the input MR for the corresponding utterance in the subsequent row.}
    \label{tab:error-detection-examples}
 	\vspace{-0.4cm}
\end{table}

All in all, \textsc{SeA-GuiDe} chooses semantically correct utterances that are fluent and adequate, except for the rare case like in the last example.


\section{Discussion}

In the previous section, we showed that \textsc{SeA-GuiDe} is highly effective at reducing semantic errors across different models and domains, and that without compromising on the generated utterances' fluency. On datasets other than E2E, it does not quite match the performance of beam search combined with our slot aligner-based reranking, but then again, the slot aligner is a hand-crafted tool with complex rules, requiring a good deal of domain knowledge, and suffering thus significantly in scalability. While these two decoding methods have a lot in common -- both being based on beam search and subsequent candidate reranking -- their difference lies in the identification of slot mentions; \textsc{SeA-GuiDe} identifies them \emph{automatically} during the decoding, utilizing the model's cross-attention weights at each step, as opposed to relying on string-matching rules post decoding, which need to be extended for any new domains.

Despite working conveniently out of the box, \textsc{SeA-GuiDe} does not come with a computational overhead caveat. Performing inference on a GPU, \textsc{SeA-GuiDe} is a mere 11--18\% slower than beam search with slot aligner-based reranking, while we observed no performance difference on a CPU (see Appendix~\ref{sec:appendix-inference-performance} for a detailed analysis).

\subsection{Limitations of \textsc{SeA-GuiDe}}

\textsc{SeA-GuiDe}'s ability to recognize slot errors is limited to missing and incorrect slot mentions, which are the most common mistakes we observed models to make on the data-to-text generation task. Duplicate slot mentions are hard to identify reliably because the decoder inherently pays attention to certain input tokens at multiple non-consecutive steps (such as in the example in Figure~\ref{subfig:attention-visualization-paraphrased}). And arbitrary hallucinations are entirely beyond the scope of this method, as there is no reason to expect cross-attention to be involved in producing input-unrelated content, at least not in a foreseeable way.

As we see in example \#4 in Table~\ref{tab:error-detection-examples}, Boolean slots occasionally give \textsc{SeA-GuiDe} a hard time, as the decoder appears not to be paying a great deal of attention to Boolean slots' values throughout the entire decoding in many cases. We plan to investigate if the performance can be improved for Boolean slots, perhaps by modifying the input format or finding a more subtle slot mention pattern.

\section{Conclusion}

We presented a novel decoding method, \textsc{SeA-GuiDe}, that makes a better use of the cross-attention component of the already complex and enormous pretrained generative LMs to achieve significantly higher semantic accuracy for data-to-text NLG, while preserving the otherwise high quality of the output text. It is an automatic method, exploiting information already present in the model, but in an interpretable way. \textsc{SeA-GuiDe} requires no training, annotation, data augmentation, or model modifications, and can thus be effortlessly used with different models and domains.


\section*{Acknowledgements}

We would like to thank the anonymous reviewers for their valuable feedback. This research was supported by NSF AI Institute Grant No.~1559735.

\bibliography{anthology,references}
\bibliographystyle{acl_natbib}

\clearpage

\appendix

\section{Appendix}

\subsection{Additional Dataset Details}

Table~\ref{tab:dataset_partitions} shows the number of examples in the training, validation and test partitions of all the datasets used in the evaluation of the \textsc{SeA-GuiDe} method.

\begin{table}
    \small
   	\centering
    \begin{tabular}{>{\centering\arraybackslash} m{0.2\linewidth} >{\centering\arraybackslash} m{0.19\linewidth} >{\centering\arraybackslash} m{0.19\linewidth} >{\centering\arraybackslash} m{0.19\linewidth}}
    	\toprule
    	& \textbf{Training}    & \textbf{Validation}    & \textbf{Test} \\
    	\midrule
    	\textbf{ViGGO}    & 5,103   & 714   & 1,083 \\
    	\textbf{E2E}    & 42,063    & 4,672 & 4,693 \\
    	\textbf{MultiWOZ}    & 55,951   & 7,286   & 7,293 \\
        \bottomrule
    \end{tabular}
    \vspace{-0.2cm}
	\caption{Overview of the dataset partitions.}
    \label{tab:dataset_partitions}
    \vspace{-0.3cm}
\end{table}

\subsection{Data Preprocessing}
\label{sec:appendix-data-preprocessing}

When preprocessing input meaning representations (MRs) before training a model or running inference, we first parse the dialogue act (DA) types, if present, and all slots and their values from the dataset-specific format into an intermediate list of slot-and-value pairs, keeping the original order. Although typically indicated in the MR differently from slots, we treat the DA type as any other slot (with the value being the DA type itself, and assigning it the name ``intent'').

Next, we rename any slots that do not have a natural-language name (e.g., ``priceRange'' to ``price range'', or ``has\_mac\_release'' to ``has Mac release''). Slot values are left untouched. We do this to take advantage of pretrained language models' ability to model the context when the input contains familiar words, as opposed to feeding it code names with underscores and no spaces.

Finally, we convert the updated intermediate list of slots and their values to a string. The `$\mid$' symbol is used for separating slot-and-value pairs from each other, while the `=' is used within each pair to separate the value from the slot name. The result for an MR from ViGGO can look as follows:

\noindent\texttt{intent = request explanation $\mid$ rating = poor $\mid$ genres = vehicular combat $\mid$ player perspective = third person}

\subsection{Model and Training Parameters}
\label{sec:appendix-model-parameters}

The pretrained models that we fine-tuned for our experiments are the PyTorch implementations in the Hugging Face's Transformers\footnote{https://huggingface.co/transformers/} package. The models' sizes are indicated in Table~\ref{tab:model-parameters}.

We trained all models using a single Nvidia RTX 2070 GPU with 8 GB of memory and CUDA version 10.2. The training parameters too are summarized in Table~\ref{tab:model-parameters}. For all models, we used the AdamW optimizer with a linear decay after 100 warm-up steps. The maximum sequence length for both training and inference was set to 128 for ViGGO and E2E, and 160 for MultiWOZ.

\begin{table}
    \small
   	\centering
    \begin{tabular}{>{\centering\arraybackslash} m{0.21\linewidth} >{\centering\arraybackslash} m{0.1\linewidth} >{\centering\arraybackslash} m{0.1\linewidth} >{\centering\arraybackslash} m{0.16\linewidth} >{\centering\arraybackslash} m{0.16\linewidth}}
    	\toprule
    	& \textbf{Layers}    & \textbf{Heads}    & \textbf{Hidden state size} & \textbf{Total parameters} \\
    	\midrule
    	\textbf{T5-small}   & 6+6   & 8     & 512   & $\approx$ 60M  \\
    	\textbf{BART-base}  & 6+6   & 12    & 768   & $\approx$ 139M \\
    	\textbf{T5-base}    & 12+12 & 12    & 768   & $\approx$ 220M \\
    	\textbf{BART-large} & 12+12 & 16    & 1024  & $\approx$ 406M \\
        \bottomrule
    \end{tabular}
    \vspace{-0.2cm}
	\caption{Overview of the model specifications.}
    \label{tab:model-parameters}
\end{table}

\begin{table}
    \small
   	\centering
    \begin{tabular}{>{\centering\arraybackslash} m{0.21\linewidth} >{\centering\arraybackslash} m{0.19\linewidth} >{\centering\arraybackslash} m{0.19\linewidth} >{\centering\arraybackslash} m{0.19\linewidth}}
    	\toprule
    	& \textbf{Batch size} & \textbf{Learning rate}    & \textbf{Epochs} \\
    	\midrule
    	\textbf{T5-small}   & 32/64/64    & \num{2e-4}  & 20/20/30  \\
    	\textbf{BART-base}  & 32/32/32    & \num{1e-5}  & 20/20/25 \\
    	\textbf{T5-base}    & 16/16/ --  & \num{3e-5}  & 20/20/ -- \\
    	\textbf{BART-large} & 16/ -- / --    & \num{4e-6}  & 20/ -- / -- \\
        \bottomrule
    \end{tabular}
    \vspace{-0.2cm}
	\caption{Overview of the training parameters used in our experiments. Batch size and the number of epochs are indicated per dataset (ViGGO/E2E/MultiWOZ).}
    \label{tab:training-parameters}
    \vspace{-0.3cm}
\end{table}

\subsection{Evaluation Metric Calculation}
\label{sec:appendix-evaluation-metrics}

The four non-SER automatic metrics that we report in our results (i.e., BLEU, METEOR, ROUGE-L, and CIDEr) are calculated using the E2E evaluation script\footnote{https://github.com/tuetschek/e2e-metrics} developed for the E2E NLG Challenge~\cite{duvsek2018findings}. We also verified that the single-reference BLEU score calculation in the E2E script corresponds to that in the SacreBLEU\footnote{https://pypi.org/project/sacrebleu/} Python package. As a result, BLEU scores calculated either way are directly comparable.

To ensure a fair comparison with the MultiWOZ baselines~\cite{peng2020few,kale2020text}, we additionally report BLEU scores calculated using the RNNLG evaluation script\footnote{https://github.com/shawnwun/RNNLG/}, which their respective authors used in their own evaluation. We denote it BLEU\textsubscript{R} in our result tables. Moreover, \citet{kale2020text} calculated SER on utterance level, rather than slot level, and that using \emph{exact} slot value matching in the utterance. We thus wrote a script to also perform this type of naive SER evaluation, in addition to our slot aligner-based SER evaluation. We report its results as SER\textsubscript{E}.

\section{Additional \textsc{SeA-GuiDe} Evaluation}

\subsection{Slot Mention Tracking Details}
\label{sec:appendix-slot-mention-tracking}

In order to be able to take advantage of the attention weight distribution patterns, the decoder needs to be aware of which input token span corresponds to which slot. To this end, we parse the input MRs on-the-fly -- which is trivial given the structured nature of MRs -- as each batch is being prepared for inference, and create a list of slot spans for each MR in the batch.\footnote{This is done on token level, and the result varies thus from model to model depending on its tokenizer.} In fact, we indicate the spans for slot names and slot values separately, and for list-values down to individual list elements, for a higher specificity. Since Boolean slot mentions are tracked by their name rather than value, we also indicate for each slot whether it is Boolean or not. This information can be provided explicitly to the data loader, otherwise it is automatically inferred from the dataset's ontology based on all the possible values for each slot.

Note that, although our data preprocessing converts DA type indications in the MRs to the same format as slots (see any of the left columns in Figure~\ref{fig:attention-visualization}), we exclude them from the slot-span lists, as they are not actual content slots to be tracked. Separator tokens (such as `$\mid$' or `=') present in the preprocessed MR are not included in the spans, and are, as a result, ignored during the slot mention tracking.

\subsection{Parameter Tuning for T5}

\begin{figure}
    \centering
    \begin{subfigure}[t]{\linewidth}
        \includegraphics[width=\linewidth]{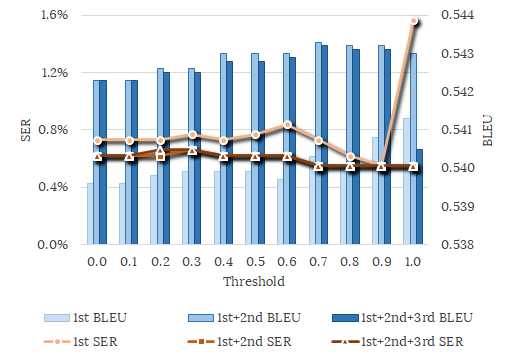}
        \caption{Threshold optimization for the 1\textsuperscript{st}~component (verbatim mentions), with the other components enabled or disabled. When enabled, the 2\textsuperscript{nd}~component's threshold was fixed at $0.3$, and that of the 3\textsuperscript{rd} at $0.1$. Note that the threshold of $1.0$ is equivalent to the 1\textsuperscript{st}~component being disabled, as attention weights are in the $[0.0, 1.0]$ range.}
        \label{subfig:thresholds-1st-component-t5}
        \vspace{0.1in}
    \end{subfigure}
    \begin{subfigure}[t]{\linewidth}
        \includegraphics[width=\linewidth]{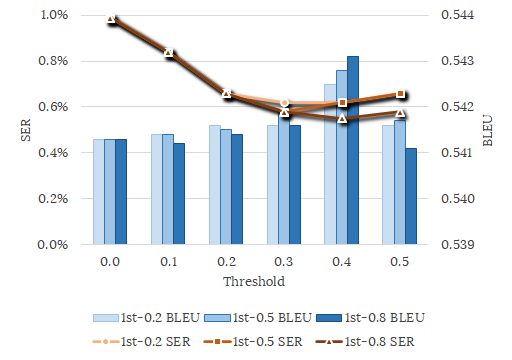}
        \caption{Threshold optimization for the 2\textsuperscript{nd}~component (paraphrased mentions), with the 1\textsuperscript{st}~component's threshold of $0.2$, $0.5$ and $0.8$, and that of the 3\textsuperscript{rd}~component fixed at $0.1$.}
        \label{subfig:thresholds-2nd-component-t5}
        \vspace{0.1in}
    \end{subfigure}
    \begin{subfigure}[t]{\linewidth}
        \includegraphics[width=\linewidth]{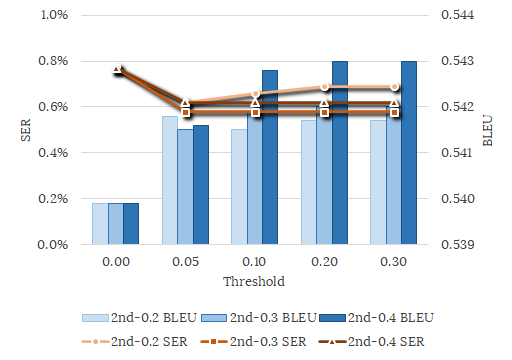}
        \caption{Threshold optimization for the 3\textsuperscript{rd}~component (unrealized mentions), with the 2\textsuperscript{nd}~component's threshold of $0.2$, $0.3$ and $0.4$, and that of the 1\textsuperscript{st}~component fixed at $0.5$.}
        \label{subfig:thresholds-3rd-component-t5}
    \end{subfigure}
    \caption{Effects of different parameter configurations of the 3 mention-tracking components on SER and BLEU of utterances generated by T5-small fine-tuned on ViGGO.}
    \label{fig:binarization-threshold-optimization-t5}
\end{figure}

When optimizing the mention-tracking components' parameters for T5-small, we observe similar trends as with BART-base (see Figure~\ref{fig:binarization-threshold-optimization-t5}). One difference is that enabling the 2\textsuperscript{nd} component not only significantly increases the BLEU score, but also lowers the SER, while the 3\textsuperscript{rd} component appears to only have a negligible effect (see Figure~\ref{subfig:thresholds-1st-component-t5}).

\subsection{Effects of Beam Size on E2E}
\label{sec:appendix-beam-size-e2e}

\begin{figure}
    \centering
    \includegraphics[width=0.95\linewidth]{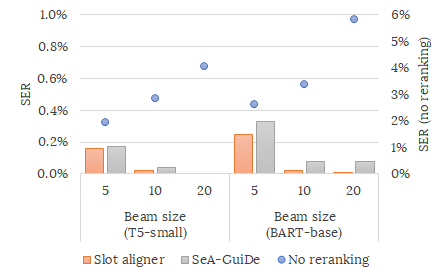}
    \vspace{-0.1in}
    \caption{The effect of different beam size on the SER using different reranking methods on the E2E dataset. With greedy search decoding, the SER is $1.60\%$ and $1.97\%$ for T5 and BART, respectively.}
    \label{fig:beam_search_e2e}
    \vspace{-0.3cm}
\end{figure}

On the E2E dataset, decoding using \textsc{SeA-GuiDe} is even more effective in reducing SER than on ViGGO. Across all beam sizes, its performance is comparable to beam search with slot aligner reranking, and there is also only a limited gain from increasing the beam size to 20 (see Figure~\ref{fig:beam_search_e2e}).

It is worth noting that, using beam search with no reranking, the SER dramatically increases with the increasing beam size. This is likely caused by the relatively heavy semantic noise in the E2E training set, resulting in more slot errors in the generated utterances the less greedy the decoding is. Some form of semantic guidance is thus all the more important for the model in this scenario.

\subsection{Inference Performance}
\label{sec:appendix-inference-performance}

In order to assess the computational overhead the \textsc{SeA-GuiDe} method introduces during inference, we measure the inference runtime of the T5-small model fine-tuned on ViGGO. For all beam search-based methods (including \textsc{SeA-GuiDe}), the beam size was set to $10$, and early stopping was enabled.

\begin{figure}[h]
    \centering
    \begin{subfigure}[t]{\linewidth}
        \includegraphics[width=\linewidth]{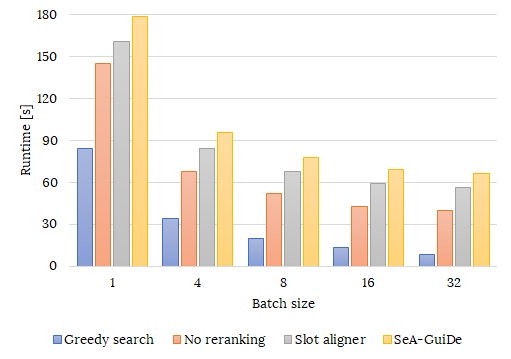}
        \caption{Inference using a GPU (RTX 2070 with 8~GB of memory).}
        \label{subfig:inference-performance-gpu}
        \vspace{0.1in}
    \end{subfigure}
    \begin{subfigure}[t]{\linewidth}
        \includegraphics[width=\linewidth]{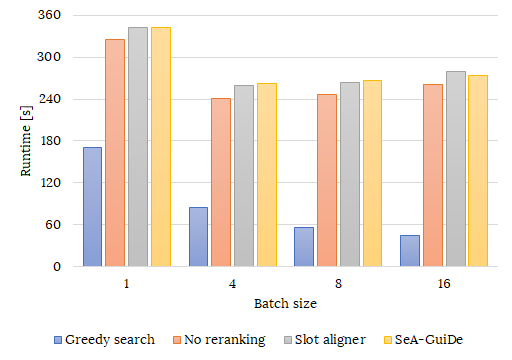}
        \caption{Inference using a CPU (8-core Ryzen 7 2700X with 32~GB of RAM).}
        \label{subfig:inference-performance-cpu}
        \vspace{0.1in}
    \end{subfigure}
    \vspace{-0.1in}
    \caption{Runtime of T5-small performing inference on the ViGGO test set using different decoding methods and batch sizes. ``No reranking'' stands for simple beam search, while ``Slot aligner'' denotes beam search with slot aligner-based reranking. Model and data loading is excluded from the runtimes.}
    \label{fig:inference-performance}
\end{figure}

The results in Figure~\ref{subfig:inference-performance-gpu} show a distinct but expected overhead across all batch sizes when running inference on a GPU. The overall increase in runtime is 11--18\% over beam search with slot aligner-based reranking, which is the method computationally most similar to \textsc{SeA-GuiDe}, as it too involves reranking on top of beam search. The slot aligner-based reranking itself adds a constant amount of 16~seconds on top of simple beam search, which corresponds to an 11-40\% increase for the range of batch sizes in the plot.

When performing the same inference on a CPU, on the other hand, the overhead \textsc{SeA-GuiDe} introduces to beam search is no greater than that of the slot aligner-based reranking (see Figure~\ref{subfig:inference-performance-cpu}). This suggests that further optimization of \textsc{SeA-GuiDe} for GPU, especially by minimizing the communication between the GPU and the CPU during the decoding, could bring the overhead of \textsc{SeA-GuiDe} inference on a GPU down to the same level as that of the slot aligner-based reranking.

Considering the large improvement in semantic accuracy the \textsc{SeA-GuiDe} method delivers in the tested models, we deem the observed computational overhead reasonable and acceptable.

\section{Slot Aligner Details}
\label{app:slot-aligner}

For the purposes of the slot aligner, we classified slots into five general categories (\emph{Boolean}, \emph{numeric}, \emph{scalar}, \emph{categorical}, and \emph{list}), covering the most common types of information MRs typically convey in data-to-text NLG. Each of these categories has its own method for extracting a slot mention from an utterance, generalized enough to be applicable across all slots in the category. This design allows for a straightforward extension of the slot aligner to a new domain, as it merely needs to be indicated which of the five categories each of the slots in the new domain belongs to. Optionally, it can be provided a simple dictionary of common alternatives for specific slot values, which tends to increase the slot aligner's performance.

Although a decreased matching accuracy -- especially for rare slot realizations -- is a trade-off for the scalable design, the slot aligner's typical application are generated model outputs, which get evaluated for semantic errors. There the slot aligner is not likely to encounter rare slot realizations frequently, if at all, due to the generalizing properties of neural NLG models. The rapid adaptability of the slot aligner to a new domain, on the other hand, is a very valuable feature.

\subsection{Boolean Slots}

Boolean slots take on binary values, such as ``yes''/``no'' or ``true''/``false''. Their realization in an utterance thus typically does not contain the actual value of the slot, but instead a mention of the slot name (e.g., ``is a family-friendly restaurant'' for \textsc{familyFriendly}[yes], or ``not supported on Mac'' for \textsc{has\_mac\_release}[no]). Therefore, extracting a Boolean slot mention boils down to the following two steps: (1) finding a word or a phrase representing the slot, and (2) verifying whether the representation is associated with a negation or not.

The first step is straightforward, and only requires a list of possible realizations for each Boolean slot. This list rarely contains more than one element, which is the ``stem'' of the slot's name (e.g., ``linux'' for ``\textsc{has\_linux\_release}''). It can thus be populated trivially for most of the new Boolean slots. And if a Boolean slot can have multiple equivalent realizations (such as ``child friendly'' or ``where kids are welcome'' for the slot \textsc{familyFriendly}), they are typically not numerous and can be listed manually. Having a list of stems (we refer to all the equivalent realizations of a slot collectively as ``slot stems''), the utterance is scanned for the presence of each of them in it. If one is found, we go to the second step. A slot mention is decided to be negative if a negation cue is found to be modifying the slot stem in the utterance, and without a contrastive cue in between. It is decided to be positive if no negation cue is present within a certain distance of the stem, or there is a contrastive cue in between (see examples in Table~\ref{tab:ex-boolean-slots-contrast}).

\begin{table}
    \small
   	\centering
    \begin{tabular}{l p{0.85\linewidth}}
    	\toprule
    	\textbf{\#1} & There's \emph{no} Linux release or multiplayer, \textbf{but} there is \underline{Mac} support. \\
        \midrule
    	\textbf{\#2} & \textbf{Though} it's \emph{not available} on Linux, it does have a \underline{Mac} release as well. \\
        \midrule
    	\textbf{\#3} & It is available on PC and Mac \textbf{but} \emph{not} Linux, and it can be found on \underline{Steam}. \\
        \bottomrule
    \end{tabular}
 	\vspace{-0.2cm}
	\caption{Examples of contrastive phrases involving Boolean slots. Underlined are the stems of the Boolean slots for which the polarity is questioned. Note that in all 3 examples the mention is positive, despite the presence of contrast and negation distractors.}
    \label{tab:ex-boolean-slots-contrast}
\end{table}

\subsection{Numeric Slots}

Slots whose value is just a number (such as \textsc{release\_year} in ViGGO, or \textsc{choice} in MultiWOZ) are in general not handled in any special way, and the value is simply matched directly in the utterance. However, there are certain numeric slot types that benefit from additional preprocessing: (1) those with a unit, and (2) years. When a numeric slot represents a year, the slot aligner generates the common abbreviated alternatives for the year (e.g., ``'97'' for the value ``1997'') that it tries to match in case the original value is not found in the utterance.

\subsection{Scalar Slots}

Similarly to Boolean slot aligning, scalar slot aligning consist of two steps. The first one is the same, i.e., finding a word or a phrase representing the slot (which we refer to as ``stem'' in this case too, in order to maintain consistency). In the second step, however, the slot aligner looks for the slot's value, or its equivalent, occurring within a reasonable distance from the slot stem. The optional soft alignment mode skips the second step as long as a slot stem is matched in the first step.

\begin{table}
    \small
   	\centering
    \begin{tabular}{>{\centering\arraybackslash} m{0.19\linewidth} >{\centering\arraybackslash} m{0.17\linewidth} p{0.48\linewidth}}
    	\toprule
    	\textbf{\textsc{customer rating} (E2E)}   & \textbf{\textsc{rating} (ViGGO)}    & \textbf{Alternative expressions} \\
        \midrule
    	low & poor  & \emph{bad, lacking, negative,\dots} \\
        \midrule
    	average & average   & \emph{decent, mediocre, okay,\dots} \\
        \midrule
    	-   & good  & \emph{fun, positive, solid,\dots} \\
        \midrule
    	high    & excellent & \emph{amazing, fantastic, great,\dots} \\
        \bottomrule
    \end{tabular}
 	\vspace{-0.2cm}
	\caption{An example of value mapping between two similar scalar slots in the restaurant and video game domains.}
    \label{tab:ex-scalar-slot-mapping}
 	\vspace{-0.3cm}
\end{table}

We assume that scalar slots, even across different domains, will often have values that can be mapped to each other, as long as they are on the same or a similar scale (see Table~\ref{tab:ex-scalar-slot-mapping}). For each scalar slot, the slot aligner refers to a corresponding dictionary for possible alternative expressions of its value. With the above assumption, it is sufficient to have one dictionary per scale, or type of scale, which can be reused for similar scalar slots in different domains. The dictionaries can be quickly populated with synonyms of the values of a given scale (see the last column in the table), and thus do not necessarily require manual additions every time the system is used with a new domain. Some alternative expressions might be suitable for scalar slots in some domains better than others, but that will not be an issue in most cases, since, being synonymous, they are not likely to cause conflicts, and the slot aligner will simply not encounter certain alternative expressions in certain domains.

\subsection{Categorical Slots}

Categorical slots can take on virtually any value. Nevertheless, for each such slot the values typically come from a limited, although possibly large, set of values. For instance, in the E2E dataset, the \textsc{food} slot has 7 possible values, such as ``Italian'' and ``Fast food'', but technically it could take on hundreds of different values representing all of the cuisines of the world. Some values can be single-word, while others can have multiple words (e.g., ``restaurant'' and ``coffee shop'' as possible values for the \textsc{eatType} slot). Due to this huge variety in possible values of categorical slots, the aligning methods need to remain very general.

Besides exact matching of the value in the utterance, the slot aligner can be instructed to perform the matching in three additional modes, besides exact, increasing its robustness while maintaining scalability. The four modes of aligning the slot with its mention work as follows:
\begin{itemize}
    \item \textbf{Exact} - slot mention is identified only if it matches (case-insensitive) the slot value verbatim;
    \item \textbf{All words} - slot mention is identified if each of the value's tokens is found in the utterance, though they can be in an arbitrary order and they can be separated by other words;
    \item \textbf{Any word} - slot mention is identified by matching any of the value's tokens in the utterance;
    \item \textbf{First word} - slot mention is identified by matching just the value's first token in the utterance.
\end{itemize}
Note that for single-word values all four modes give the same result. The three non-exact modes offer different approaches to soft alignment for categorical slots. The choice may depend on the particular slot, and the mode can thus be specified for each slot separately, while by default the slot aligner operates in the exact-matching mode.

Similarly to Boolean and scalar slots, the slot aligner can search for alternative expressions of a value, if provided in the corresponding dictionary. The alternative matching is, however, more flexible here, as the alternatives in the dictionary can be multi-part, in which case the slot aligner tries to match all the parts (words/tokens/phrases) provided in the form of a list.

\subsection{List Slots}

A list slot is similar to a categorical slot, the only difference being that it can have multiple individual items in its value. Two instances of a list slot, namely \textsc{genres} and \textsc{platforms}, can be seen in the example from the ViGGO dataset in Table~\ref{tab:ex-list-slot-alignment}.

\begin{table}
    \small
   	\centering
   	\def\arraystretch{1.5}
    \begin{tabular}{p{0.94\linewidth}}
    	\hline
    	\rowcolor{light-gray}\textbf{MR} \\
        \hline
    	\emph{inform}(\textsc{name} [\textbf{BioShock}], \textsc{developer} [\textbf{2K Boston}], \textcolor{blue}{\textsc{genres} [\textbf{action-adventure, role-playing, shooter}]}, \textsc{has\_mul\-ti\-player} [\textbf{no}], \textcolor{orange}{\textsc{platforms} [\textbf{PlayStation, Xbox, PC}]}, \textsc{has\_lin\-ux\_re\-lease} [\textbf{no}], \textsc{has\_mac\_re\-lease} [\textbf{yes}]) \\
        \hline
    	\rowcolor{light-gray}\textbf{Reference utterance} \\
        \hline
    	Developed by \textbf{2K Boston}, \textbf{BioShock} is a \textbf{single-player} \textcolor{blue}{\textbf{shooter}} game that will have you \textcolor{blue}{\textbf{role-playing}} through a well constructed \textcolor{blue}{\textbf{action-adventure}} narrative. It is \textbf{available for} \textcolor{orange}{\textbf{PlayStation}}, \textcolor{orange}{\textbf{Xbox}}, \textbf{Mac} and \textcolor{orange}{\textbf{PC}}, but is \textbf{not available for Linux}. \\
        \hline
    	\rowcolor{light-gray}\textbf{Slot alignment} \\
        \hline
    	(\textbf{13:} \textsc{developer}) (\textbf{25:} \textsc{name}) (\textbf{39:} \textsc{has\_multiplayer}) (\textbf{53:} \textsc{genres}) (\textbf{174:} \textsc{platforms}) (\textbf{191:} \textsc{has\_mac\_re-lease}) (\textbf{228:} \textsc{has\_linux\_release}) \\
        \hline
    \end{tabular}
	\caption{An example from ViGGO that involves list slots. Notice how the individual value item mentions can be scattered across an entire sentence in a natural way. The bottom section indicates the slot mention positions determined by the slot aligner, given as the number of characters from the beginning of the utterance.}
    \label{tab:ex-list-slot-alignment}
 	\vspace{-0.3cm}
\end{table}

The aligning procedure for list slots thus heavily relies on that of categorical slots. In order to align a list slot with the corresponding utterance, the slot aligner first parses the individual items in the slot's value. It then iterates over all of them and performs the categorical slot alignment, as described in the previous section, with each individual item. Considering the items can be scattered over multiple sentences, the slot aligner considers the position of the leftmost mention of an item as the position of the corresponding list slot.

\end{document}